\documentclass{article}
\usepackage[utf8]{inputenc}
\usepackage[T1]{fontenc}
\usepackage{microtype}
\usepackage{graphicx}
\usepackage{subcaption}
\usepackage{booktabs} 
\usepackage{cite}      
\usepackage{tabularx}
\usepackage[table]{xcolor}   
\usepackage{amsmath}
\usepackage{hyperref}
\usepackage[margin=1in]{geometry}        
\usepackage{times}                       
\usepackage{setspace}                    
\usepackage{lipsum}
\usepackage{booktabs, subcaption, makecell}
\usepackage{pgfplots}
\usepackage{authblk}
\pgfplotsset{compat=1.18}

\title{Data-Driven Fire-Zone Segmentation for Improved Short-Term Wildfire Prediction}

\author[1]{Nicolas Caron}
\author[1]{Hassan Noura}
\author[1]{Christophe Guyeux}
\author[2]{Benjamin Aynes}

\affil[1]{Université Maris et Louis Pasteur, FEMTO-ST, Besançon, France}
\affil[2]{SAD Marketing, Lille, France}

\date{}

\begin{document}
\maketitle

\begin{abstract}
	Wildfire prediction models typically discretize study areas into uniform grids, ignoring the heterogeneous spatial distribution of ignitions. We challenge this paradigm by showing that \emph{how} data is discretized matters more than \emph{which} model is used. We propose an unsupervised fire-zone segmentation algorithm combining watershed detection with K-means clustering to define prediction units directly from historical fire patterns. Experiments across six French departments and six forecasting models show that fire-zone segmentation consistently outperforms grid-based approaches, with mean IoU improvements of +3--6\% depending on spatial scale. The method is computationally lightweight ($<$10s per configuration) and fully parallelizable. Our results demonstrate that optimizing spatial discretization yields significant, reproducible performance gains for short-term wildfire forecasting. Supplementary materials are available~\href{https://drive.google.com/drive/folders/1-QOrC0Bk0PFfP3KSfrjJSpfF2M9tAMJk?usp=sharing}{\color{blue}{here}}, and code will be released on GitHub.
\end{abstract}


\section{Introduction}
Wildfires affect every continent, arising from climatic, ecological, and anthropogenic factors. Globally, nearly half a million events are reported annually, with over 90\% attributed to human activity~\cite{articleHumancaused}. The economic and humanitarian burden is substantial: suppression costs, infrastructure damage, and public-health impacts routinely exceed billions of US dollars per fire season~\cite{inventions7010015}. Early warning systems integrating IoT sensor networks with artificial intelligence (AI) have shown considerable potential for improving detection and resource allocation~\cite{GIANNAKIDOU2024101171}.

\subsection{Problem formulation}
This work addresses the detection and spatial segmentation of fire-prone zones as a preprocessing step for wildfire prediction. Specifically, we investigate whether partitioning the study area into semantically meaningful fire zones—rather than relying on arbitrary uniform grids—can improve the accuracy of short-term wildfire forecasts generated by AI models.

\subsection{State of the art}

Artificial intelligence and Internet-of-Things technologies already support wildfire detection and preparedness, yet predictive performance depends critically on how space is discretized~\cite{GIANNAKIDOU2024101171}. CNN and ULSTM variants have been explored at kilometre-scale resolutions in California and the Mediterranean~\cite{WildfireDangerPrediction, kondylatos2023mesogeosmultipurposedatasetdatadriven}; however, results at such fine scales are difficult to interpret due to the overwhelming prevalence of fire-free cells, which necessitates undersampling and biases validation metrics. Several studies~\cite{vilar2010model, PartIImprovingWildfireOccurrencePredictionforCONUSUsingDeepLearningandFireWeatherVariables} report declining prediction quality when targeting very fine resolution, calling into question the reliability of pixel-level forecasts. Furthermore, aggregated information---such as total fire counts or cumulative burned area---is lost at this granularity. Coarser spatial resolutions can mitigate these issues, but the question of how to delineate zones then arises. Contemporary work typically relies on uniform grids: Michail et al.~\cite{michail2025firecastnetearthasagraphseasonalprediction} coupled GraphCast~\cite{lam2023graphcastlearningskillfulmediumrange} with a temporal encoder on a $0.25^{\circ}$ grid; Chen et al.~\cite{rs14174391} partitioned Portugal into ten cells to predict burned area; and Koh et al.~\cite{koh2023gradient} analyzed U.S.\ risk on a $0.5^{\circ}$ grid.

Recent work has increasingly framed wildfire mapping as a \emph{supervised} image-segmentation task using deep learning architectures such as U-Net, Vision Transformers (ViT), and attention-based models~\cite{huot2022deep, ban2020unet, fire7040133}. These methods excel at detecting active fires or delineating burned areas from multispectral satellite imagery when pixel-level ground-truth labels are available. However, they address a \emph{different problem}: detecting or segmenting fires \emph{after} ignition, rather than defining prediction zones \emph{before} forecasting. No public dataset provides pre-labeled fire-prone zones suitable for training such models in a predictive context. Our approach is therefore \emph{unsupervised}: it derives prediction units directly from historical ignition distributions without requiring segmentation labels, making it complementary to---rather than competing with---supervised deep-learning pipelines.

A key limitation of the state of the art in daily wildfire prediction is its reliance on regular grid-based segmentation, which disregards the complex, irregular spatial distribution of wildfires. Such grids introduce noise into model training (water bodies, unreliable historical encoding, etc.), potentially degrading predictive performance. In this article, we address this gap by proposing a fire-zone segmentation algorithm that leverages image-processing techniques to define semantically meaningful prediction zones. Our results demonstrate clear improvements over conventional grid-based approaches, highlighting the importance of spatially aware segmentation for wildfire prediction. Table~\ref{tab:related_work} summarizes key differences between our approach and recent related work.

\begin{table}[htbp]
	\centering
	\caption{Comparison of spatial discretization approaches for wildfire prediction.}
	\footnotesize
	\setlength{\tabcolsep}{3pt}
	\renewcommand{\arraystretch}{1.15}
	\begin{tabular}{lcccc}
		\toprule
		\textbf{Study}                                                       & \textbf{Resolution}  & \textbf{Segmentation} & \textbf{Supervised} & \textbf{Task} \\
		\midrule
		Michail~\cite{michail2025firecastnetearthasagraphseasonalprediction} & $0.25^{\circ}$       & Grid                  & N/A                 & Forecast      \\
		Chen~\cite{rs14174391}                                               & 10 cells             & Grid                  & N/A                 & Forecast      \\
		Koh~\cite{koh2023gradient}                                           & $0.5^{\circ}$        & Grid                  & N/A                 & Risk          \\
		Huot~\cite{huot2022deep}                                             & Pixel                & None                  & Yes                 & Detection     \\
		Ban~\cite{ban2020unet}                                               & Pixel                & None                  & Yes                 & Detection     \\
		\midrule
		\textbf{Ours}                                                        & $0.2$--$0.4^{\circ}$ & \textbf{Fire-zone}    & \textbf{No}         & Forecast      \\
		\bottomrule
	\end{tabular}
	\label{tab:related_work}
\end{table}

\subsection{Contributions}
\begin{figure*}[htbp]
	\vskip 0.2in
	\begin{center}
		\resizebox{\textwidth}{!}{\includegraphics{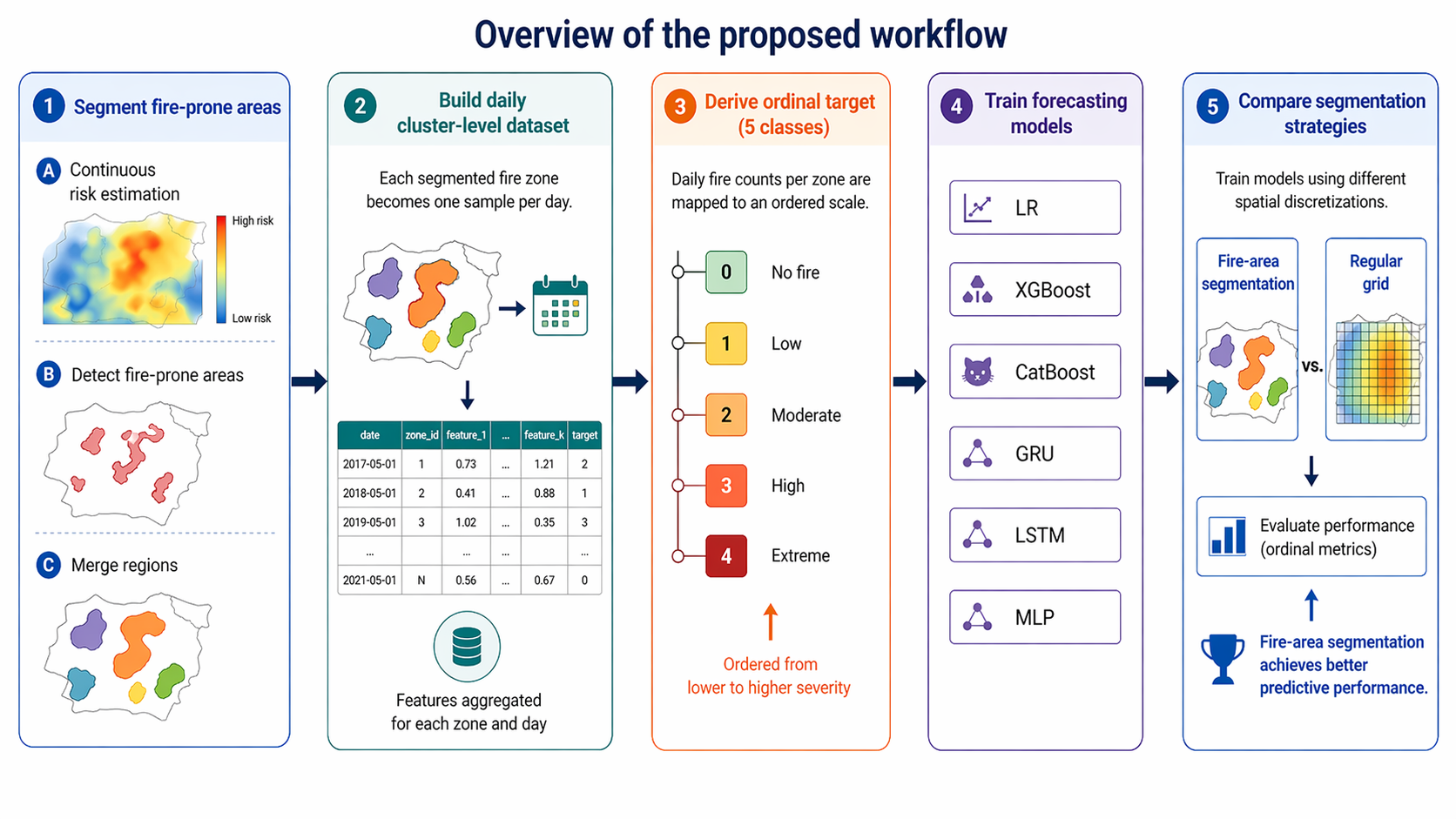}}
		\caption{Process pipeline. First, the algorithm detects and segments fire-affected zones in which risk will be forecast. Next, a daily dataset is built with each fire cluster treated as an individual sample. Finally, each model is trained on both the fire-area segmentation and a conventional regular grid to demonstrate the advantages of our method.}
		\label{fig:pip}
	\end{center}
	\vskip -0.2in
\end{figure*}

This article addresses a central issue in wildfire modeling: the definition of prediction units (grid cells vs.\ fire zones) conditions the model's capacity to learn ignition patterns. We demonstrate that grid-based spatial discretization constitutes a major bottleneck in wildfire prediction---it dilutes the ignition signal and introduces spatial noise. To overcome this limitation, we introduce a fire-area segmentation algorithm that combines watershed detection with clustering to align prediction zones with historical ignition patterns and thereby improve short-term forecasting. Models trained on this segmentation consistently outperform those using a standard grid across multiple spatial scales. While this partitioning approach enhances predictive performance, we do not claim optimality, and no systematic comparison with alternative segmentation strategies has yet been conducted. Moreover, no public dataset provides a validated spatial segmentation to support standardized evaluation. This work opens a different research perspective in wildfire prediction---one that does not seek new models but rather optimizes how the dataset itself is constructed. Figure~\ref{fig:pip} illustrates the pipeline developed in this study.

\section{Proposed Method: Fire-Area Segmentation}\label{Segmentation}

Conventional grid-based discretization methods that partition the study area into uniform cells are suboptimal because fire ignitions follow complex, non-uniform spatial patterns. Accordingly, our approach prioritizes cluster size, ensuring adaptability across departments and scales. Since no public dataset exists specifically for \textit{pre-fire} fire-zone segmentation, we rely on unsupervised segmentation algorithms. The proposed pipeline segments fire-prone areas prior to any predictive modeling. Our method comprises three main stages: (A)~generating a continuous 3D signal of fire risk, (B)~detecting fire-prone areas, and (C)~merging fire areas to match a specified target size.

\noindent\textbf{Algorithmic design choices.} We selected watershed segmentation combined with K-means clustering over alternative approaches such as SLIC superpixels~\cite{achanta2012slic}, Mean-Shift~\cite{comaniciu2002meanshift}, or DBSCAN for the following reasons. First, watershed naturally detects ``basins'' in the density surface generated from historical ignitions, producing zones that follow the topography of fire risk rather than imposing arbitrary geometric shapes. Second, K-means provides explicit control over the \emph{number} of merged regions via the \texttt{intensity\_levels} parameter, which is essential to match operational constraints (departmental boundaries, target zone sizes). Third, both algorithms are computationally lightweight and deterministic given a fixed random seed, enabling reproducible experiments across multiple configurations. SLIC, while effective for image oversegmentation, does not adapt well to sparse, irregularly distributed point data; Mean-Shift is sensitive to bandwidth selection and can produce unpredictable cluster counts; DBSCAN struggles when density varies significantly across the study area. Our two-stage approach (watershed + K-means merging) balances spatial fidelity with controllable granularity.

\begin{figure*}[htbp]
	\centering
	\vskip -0.2in
	\includegraphics[scale=0.25]{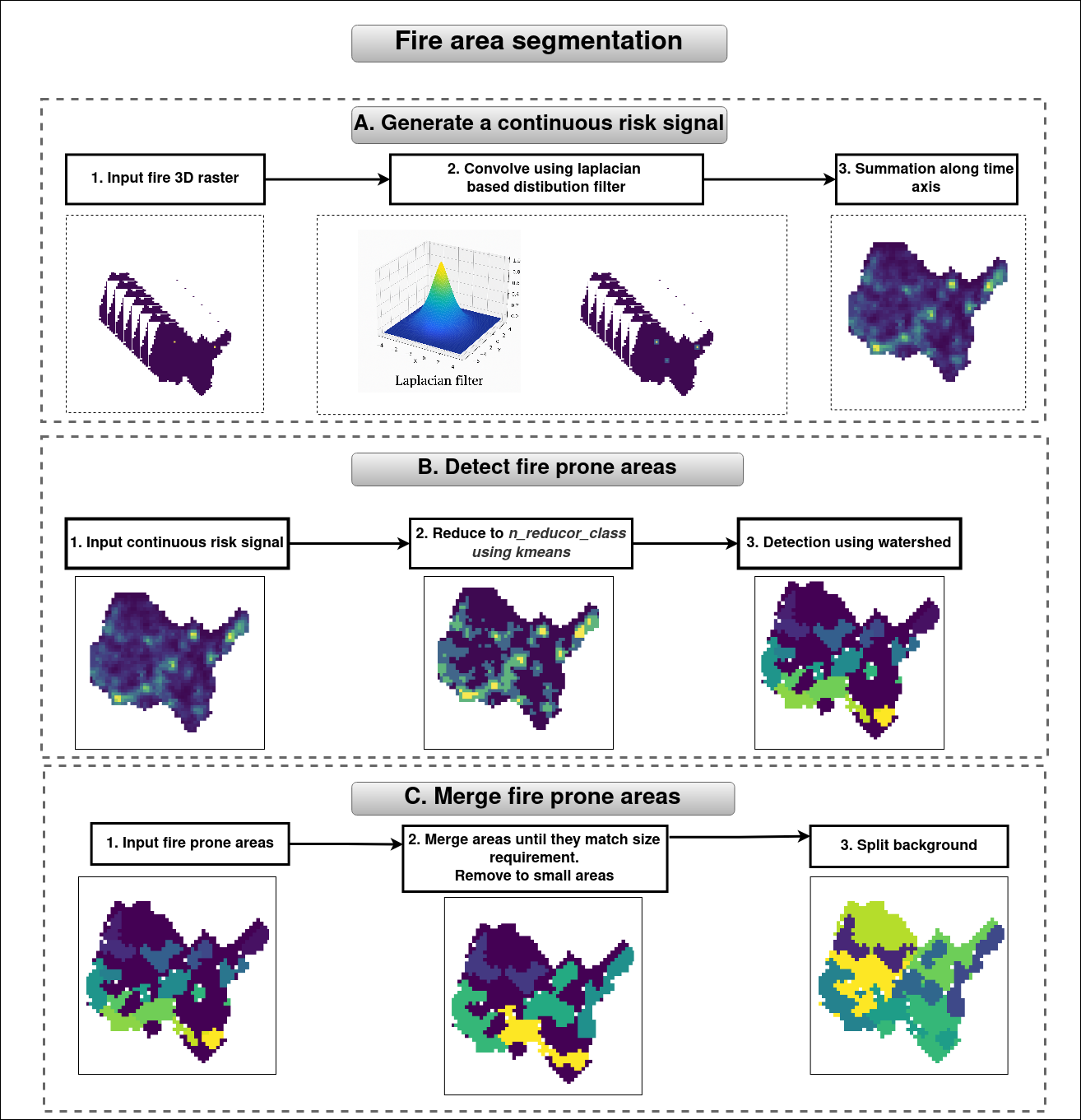}
	\caption{Detailed steps of the image-processing method implemented in this article. White pixels represent hexagons that were incorrectly rasterized at 2 km resolution and were removed.}
	\label{fig:continuous}
	\vskip -0.2in
\end{figure*}

Figure~\ref{fig:continuous} illustrates the detailed steps of the image-processing method. The objective of this algorithm is to construct optimal prediction zones---i.e., zones in which the number of fires can be predicted---for a specified zone size, based on the locations of known fires. The same algorithm can be applied to risk locations provided by upstream AI models when prediction zones are not known in advance.

The method takes three input parameters:
\begin{enumerate}
	\item \textbf{Scale} (s): Determines the size of the areas in degrees. Smaller scale values result in more zones, while larger values reduce their number by merging areas. The number of pixels (size) corresponds to the number of 2 km resolution pixels within a square of $\text{scale} \times \text{scale}$ degrees.
	\item \textbf{intensity\_levels} (r): Controls the number of intensity thresholds for fire regions. Higher values generate larger intensity zones, while lower values focus on localizing the highest-risk areas.
	\item \textbf{max\_dilations} (a): Specifies the maximum number of dilation iterations applied to merge undersized zones. Higher values produce more connected regions but may introduce noise.
	\item \textbf{tol} (t): Defines a tolerance on the region size. Since the data are organized by department, it is difficult to obtain regions that exactly match the desired size (pixel-perfect), as fire zones do not necessarily follow particular sizes and may vary across departments. However, some may be sufficiently close. This parameter allows defining a maximum and minimum size such that $\text{size} \pm \text{size} \times \text{tol}$ with size being the number of pixels of 2km resolution in a $\text{scale} \pm \text{scale}$ square.
\end{enumerate}

\subsection{Continuous risk signal}
The raw fire occurrence raster is too sparse and discrete to be directly used for spatial segmentation, as ignition points are isolated and do not form continuous regions. To enable the watershed algorithm to detect coherent fire-prone areas, we first transform this discrete signal into a continuous spatial risk surface.
Risk is first estimated for each pixel using a Laplacian distribution-based filter applied to a 3D raster of fire occurrences. It smooths the signal based on the average duration of fire sequences within a 20 km radius (computed seasonally and regionally). Summing along the temporal axis yields cumulative risk.

\subsection{Detect fire-prone area}
This step reduces noise and helps identify the main fire areas.
The cumulative risk is clustered into \textit{n\_reducor\_class\_class} groups using K-means. The \textit{n\_reducor\_class\_class} parameter is used to select the intensity of the fire-prone areas detected by the watershed algorithm.

\subsection{Merge fire-prone areas}

The purpose of this step is to ensure that each detected fire-prone region matches a target size corresponding to the spatial scale at which predictions will be made. In other words, the algorithm does not only detect fire-prone areas, but reshapes them so that they become valid prediction units.
The pseudo code, Python code, and a detailed figure are available in the supplementary materials. Figure~\ref{fig:merge} shows a simple diagram of the merge algorithm used in this article.

Zones smaller than the minimum size are merged with neighboring zones. The algorithm selects the largest neighboring zone that yields a cluster of the required size. When no valid neighbor is found, areas are expanded (dilated) to reach more distant neighbors. Undersized zones persisting after \textit{max\_dilations} iterations are removed, while oversized zones are eroded. After each merge, the algorithm reruns until every cluster is validated. If none of the zones meet the criteria, the output is an image containing only the background (i.e., no segmented regions). Background pixels (i.e., non-fire pixels) are recursively split until all clusters reach the required size ($\text{size} \pm \text{size} \times \text{tol}$).

The output of the algorithm is a labeled raster in which each pixel belongs to a fire-zone cluster that defines a prediction unit. On average, the algorithm completes a segmentation in 40 seconds, including I/O operations. This benchmark was performed on a Dell Precision 7780 with 32\,GB of RAM and a 13th-generation Intel Core i7-13850HX (28 cores). Computation time scales with the size of the study area. Since the algorithm does not explicitly account for the size of the original study region, no modifications are required to apply it at larger scales (e.g., an entire country). Applying the algorithm at finer resolution (e.g., $1 \times 1$\,km) likewise requires no changes, because the number of pixels within a zone---not the pixel resolution itself---is the relevant quantity.


\begin{figure*}[htbp]
	\vskip 0.2in
	\begin{center}
		\centerline{\includegraphics[scale=0.12]{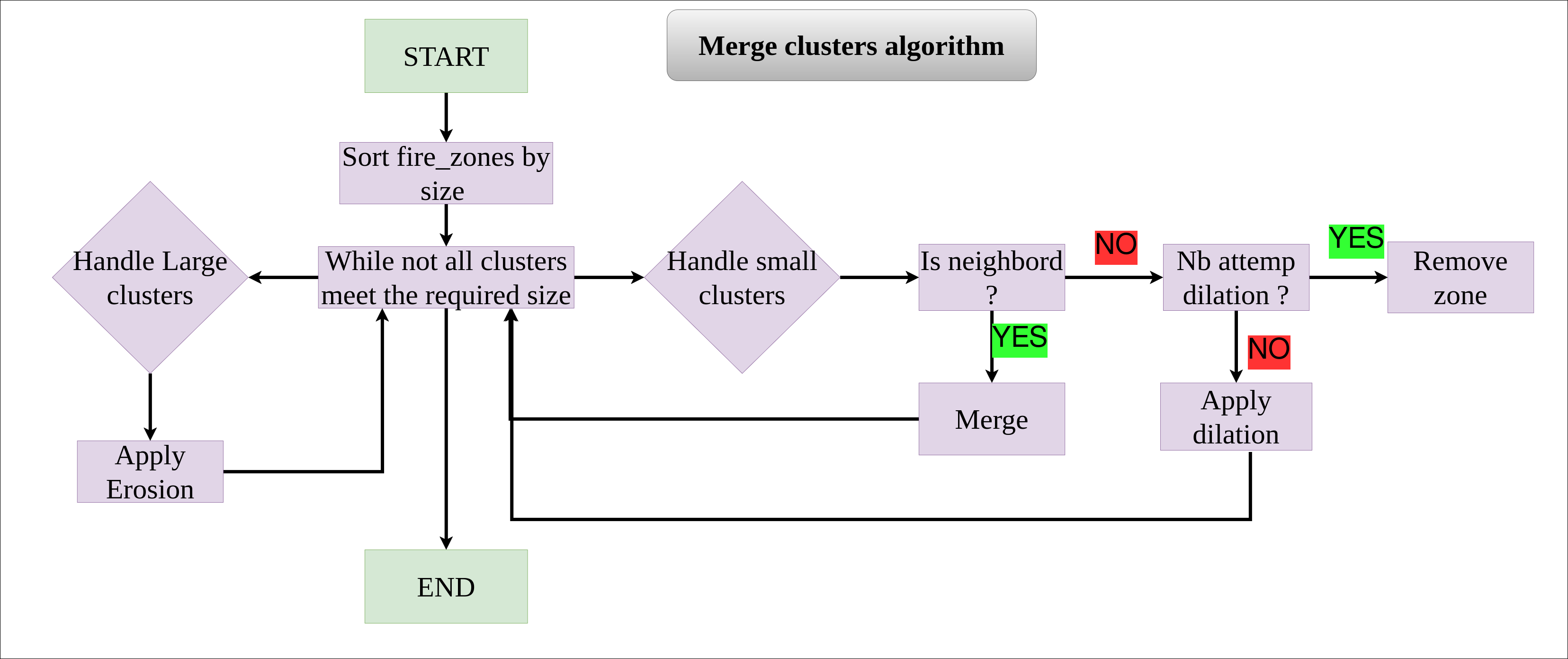}}
		\caption{Simple scheme of the merging cluster algorithm used in this article.}
		\label{fig:merge}
	\end{center}
	\vskip -0.2in
\end{figure*}




The grid-based segmentation was obtained by recursively splitting each department in two until clusters of size $\text{size} \pm \text{tol} \times \text{size}$ were produced. The splitting was performed using a 2-cluster K-Means algorithm. A split is done only when doing so makes the resulting cluster sizes closer to the target size than the original cluster.

\section{Datasets}\label{features}

This section describes the two datasets used to validate our method. Both are organized by French \emph{departments}---administrative divisions situated between the regional and municipal levels.

\subsection{Fire data}

\noindent\textbf{Firefighter interventions.}  This dataset compiles wildfires that required firefighter intervention across four French departments: Ain (01), Doubs (25), Rh\^one (69), and Yvelines (78). These regions have recently experienced increasingly severe fire seasons with a rising number of recorded incidents; however, they have received little research attention. Most fires result from human activity, although their precise causes remain unknown. The recorded location corresponds to the position from which the fire was reported. Data span from 2017 to 2024, except for Rh\^one, which ends in 2022.

\noindent\textbf{BDIFF.} \href{https://bdiff.agriculture.gouv.fr/}{The Forest Fire database} has centralized data on forest fires in France since 2006. Each fire is recorded based on the municipality in which it occurred. We selected the Bouches-du-Rh\^one and H\'erault departments, which are well known for their vulnerability to fires. Note that forest fires may not fully represent a department's fire risk, as other types of fires (e.g., wildland or natural-area fires) are not included. To the best of our knowledge, no prior study has made this distinction. Data span from 2017 to 2023.

These datasets are well suited to evaluate spatial discretization because fire locations are imprecise and heterogeneous.

\subsection{Features and Target}

The features and the target are loaded as was done in the article by Caron et al.~\cite{caron2025localizedforestriskprediction}. Wildfire forecasting is treated as an ordinal multi-class task. For every department, a five-level occurrence label is created: days with no fires are placed in class 0, while all positive instances are grouped by K-means into four ordered categories—Normal, Medium, High, and Extreme. This scheme emphasizes typical fire-activity patterns instead of relying on absolute counts. The ordinal target is unrelated to the segmentation process and simply reflects how wildfire occurrence is modeled in this study.

The features computed for training the models are grouped into six categories, as shown in Table~\ref{tab:variables}: Meteorological, Topographic, Socio-Economic, Air Quality, Hydrological, and Historical. Feature processing is the same method as in the original article. We add \textit{Air quality} and \textit{Hydrological} variables for \textit{Firefighter} dataset. The integration follows the same scheme: features are transformed into a 3D raster with a resolution of 2 km, aggregated using average, maximum, and minimum. Features with low variance or high correlation (measured with Pearson, Spearman, and Kendall methods, keeping those with the highest variance) are removed. The 2 km resolution is not a requirement of the segmentation algorithm. It was chosen as a practical compromise between spatial precision and the uncertainty in reported fire locations, ensuring that each pixel reasonably covers the area where a fire likely occurred.

The datasets are split into training (2017--2021), validation (2022), and testing (2023) subsets. All preprocessing was performed on the training set and then applied to the remaining data. The \textbf{Firefighters} and \textbf{BDIFF} datasets are treated separately. Table~\ref{tab:Dataset} presents the distribution of fires across departments. Each row consists of the aggregated features and the target risk for a specific cluster on a specific day. The segmentation was performed using the training set only.

\begin{table*}[htbp]
	\begin{center}
		\caption{Summary of features used in this study. '-' means the same as above.}
		\vskip 0.15in
		\label{tab:variables}
		\resizebox{\textwidth}{!}{\begin{tabular}{lll|lll}
				\toprule
				\textbf{Variables}                          & \textbf{Frequency}                       & \textbf{Source}                                                    & \textbf{Variables}           & \textbf{Frequency} & \textbf{Source}                                                                                \\
				\midrule
				\multicolumn{3}{l}{\textbf{Meteorological}} & \multicolumn{3}{l}{\textbf{Topographic}}                                                                                                                                                                                                                           \\
				\midrule
				Temperature                                 & 12h, 16h                                 & \href{https://dev.meteostat.net/python/}{Meteostat}                & Elevation                    & Static             & \href{https://www.ign.fr/}{IGN}                                                                \\
				Dew Point                                   & -                                        & -                                                                  & Forest landcover             & -                  & -                                                                                              \\
				Precipitation                               & -                                        & -                                                                  & Landcover                    & -                  & \href{https://geoservices.ign.fr/cosia}{Cosia}                                                 \\
				Wind Direction                              & -                                        & -                                                                  & NDVI, NDSI, NDMI, NDBI, NDWI & 7 days             & \href{https://developers.google.com/earth-engine/datasets/catalog/landsat}{GEE  (landsat 1+2)} \\
				Wind Speed                                  & -                                        & -                                                                  & Swelling-shrinking of clays  & -                  & -                                                                                              \\
				Precipitation in Last 24 hours              & -                                        & -                                                                  &                              &                    &                                                                                                \\
				Snow height                                 & -                                        & -                                                                  &                              &                    &                                                                                                \\
				Sum of last 7 days rain drop                & -                                        & -                                                                  &                              &                    &                                                                                                \\
				Day since last rain                         & 12h                                      & -                                                                  &                              &                    &                                                                                                \\
				Nesterov                                    & -                                        & \href{https://github.com/steidani/FireDanger}{firedanger}          &                              &                    &                                                                                                \\
				Munger                                      & -                                        & -                                                                  &                              &                    &                                                                                                \\
				KBDI                                        & -                                        & -                                                                  &                              &                    &                                                                                                \\
				Angstroem                                   & -                                        & -                                                                  &                              &                    &                                                                                                \\
				BUI, ISI, FFMC, DMC, FWI,                   & -                                        & -                                                                  &                              &                    &                                                                                                \\
				Daily severity rating                       & -                                        & -                                                                  &                              &                    &                                                                                                \\
				Precipitation Index last 3, 5, and 9 days   & -                                        & Calculated                                                         &                              &                    &                                                                                                \\
				\midrule
				\multicolumn{3}{l}{\textbf{Socio-Economic}} & \multicolumn{3}{l}{\textbf{Historical}}                                                                                                                                                                                                                            \\

				\midrule
				Highway                                     & Static                                   & OSMNX~\cite{BOEING2017126}                                         & Past risk                    & Daily              & Calculated                                                                                     \\
				Population                                  & -                                        & \href{https://www.kontur.io/portfolio/population-dataset/}{Kontur} & Past risk burned area        & -                  & -                                                                                              \\
				Calendar                                    & Daily                                    &                                                                    & Cluster                      & Static             & -                                                                                              \\
				                                            &                                          &                                                                    & Department                   & -                  & -                                                                                              \\
				\midrule
				\multicolumn{6}{l}{\textbf{Air Quality (Only for Firefighter dataset)}}                                                                                                                                                                                                                                          \\
				\midrule
				O3                                          & Daily                                    & \href{https://www.geodair.fr/}{Geodair}                            & NO2                          & -                  & -                                                                                              \\
				PM10                                        & -                                        & -                                                                  & PM25                         & -                  & -                                                                                              \\
				\bottomrule
				\multicolumn{6}{l}{\textbf{Hydrology (Only for Firefighter dataset)}}                                                                                                                                                                                                                                            \\
				\midrule
				River height                                & Daily                                    & \href{https://www.vigicrues.gouv.fr/}{vigicrues}                   & Groundwater depth            & Daily              & \href{https://hubeau.eaufrance.fr/}{eaufrance}                                                 \\
				\bottomrule
			\end{tabular}
		}
	\end{center}
	\vskip -0.1in
\end{table*}

\begin{table*}[htbp]
	\begin{center}
		\caption{Statistics in the regions of our dataset.}
		\vskip 0.15in
		\resizebox{\textwidth}{!}{\begin{tabular}{|l|c|c|c|c|c|c|}
				\hline
				              & Ain (01)   & Doubs (25) & Yvelines (78) & Rhone (69) & Bouche du Rhone (13) & Herault (34) \\
				\hline
				Started date  & 2018-01-01 & 2012-09-01 & 2017-01-01    & 2018-01-01 & 2017-01-01           & 2017-01-01   \\
				Ended date    & 2024-06-29 & 2024-06-29 & 2024-06-29    & 2022-12-31 & 2023-12-31           & 2023-12-31   \\
				\hline
				Wildfire fire & 2972       & 2980       & 1906          & 1324       & 1816                 & 1025         \\
				\hline
			\end{tabular}
		}
		\label{tab:Dataset}
	\end{center}
	\vskip -0.1in
\end{table*}

\section{Training}
This section presents the training scheme employed to validate the segmentation methods. For each segmentation approach (watershed-based or grid-based), every model is trained to predict daily fire risk within each cluster.

Table~\ref{tab:models} lists the models evaluated in this study. Detailed hyperparameter values for each model are provided in the supplementary materials, along with the Python code.

The objective of this study is to demonstrate the benefits of segmenting fire-affected areas prior to short-term prediction. To that end, we selected fast-to-train models so as to keep the hyperparameter search within an acceptable time budget---on the order of a few days---using five AMD EPYC 7513 (Zen 3) CPUs with 256\,GB of RAM. This choice is further justified by recent benchmarks showing that gradient-boosted decision trees (XGBoost, CatBoost) remain highly competitive with---and often outperform---deep learning models on tabular data~\cite{grinsztajn2022why}. More computationally demanding architectures such as Transformers or CNNs could be evaluated in a subsequent phase once the segmentation parameters have been finalized and the concept validated.

Because the segmentation algorithm can produce different clusters on each run, optimizing these clusters is essential. A grid-search strategy over \textit{max\_dilations} and \textit{intensity\_levels} was used to generate a large number of candidate partitions for training each model. For each segmentation method (grid-based or fire-area-based), a new dataset is constructed in which each sample corresponds to one zone on one day. All models are then trained and evaluated under identical conditions. Consequently, the comparison is between two different ways of constructing the dataset rather than between models.
To this end, we firstly fixed the parameter $\mathit{tol}$ at $0.3$, the
highest tolerance value for which no overlap is observed in the cluster‐size
distribution across scales, with a step of 0.1° (see table~\ref{tab:cluster-size-range}). The search grid was then specified as
\[
	\begin{aligned}
		\mathit{max\_dilations}      \in \{1,2,3,4,5\}, \\
		\mathit{intensity\_levels} \in \{2,3,4,5,6\}.
	\end{aligned}
\]

Analysis were done on three different scales (in degrees):
\[
	\mathit{scales} \in \{0.2°, 0.3°, 0.4°\},
\]

By comparing results across a regular parameter grid, we identified the combination that maximizes predictive performance---separately for each spatial scale and forecasting horizon. All experiments used a fixed random seed of 42 to ensure reproducibility.

We observe that the optimal parameters change across all configurations of models, departments, scales, and forecast horizons. Optimal values of $max\_dilations$ and $intensity\_levels$ are provided in the supplementary materials.

To address class imbalance, we tested different proportions of zero-class samples (from 0.05 to 1.0) and selected the proportion yielding the best validation score. Across most configurations, a proportion of 0.3--0.5 proved optimal, balancing signal retention with noise reduction.

\begin{table}[htbp]
	\centering
	\caption{Range of cluster sizes at each scale studied with a tolerance of 0.3}
	\label{tab:cluster-size-range}
	\resizebox{0.8\columnwidth}{!}{\begin{tabular}{lccc}
			\toprule
			\textbf{Scale} & \textbf{Size (avg.)} & \textbf{Minimum size} & \textbf{Maximum size} \\
			\midrule
			0.2            & 169 px               & 84 px                 & 157 px                \\
			0.3            & 256 px               & 179 px                & 332 px                \\
			0.4            & 677 px               & 338 px                & 629 px                \\
			\bottomrule
		\end{tabular}
	}
\end{table}

\begin{table}[htbp]
	\centering
	\caption{List of Models Used (LR = Logistic Regression)}
	\resizebox{0.4\textwidth}{!}{%
		\begin{tabular}{|l|l|p{8cm}|}
			\hline
			\textbf{Model}           & \textbf{Type}             \\
			\hline
			LR (Logistic Regression) & Linear model              \\
			XGBoost                  & Decision tree (Boosting)  \\
			CatBoost                 & Decision tree (Boosting)  \\
			GRU                      & Recurrent neural network  \\
			LSTM                     & Recurrent neural network. \\
			MLP                      & Multi Layer Perceptron    \\
			\hline
		\end{tabular}
	}
	\label{tab:models}
\end{table}

\section{Results}
The models' performance is evaluated using the IoU metric which compute the intersection over union between the predicted and the real signal. It is well-suited for multiclass wildfire prediction as it accounts for class uncertainty and preserves class ordinality—predicting class 1 instead of 4 is penalized less than predicting 0~\cite{ai6100253}. However, IoU can underestimate performance on rare classes (i.e., high-intensity fires), which motivates the undersampling strategy described in Section~III-D. The binary F1 score is shown in the supplementary materials and leads to a similar conclusion.

\subsection{Firefighter interventions}

Table~\ref{tab:performance_firemen} compares the performance between grid and fire area segmentation in the \textbf{Firefighter} dataset. Table~\ref{tab:mean_gain_firemen_leadtime} shows the average IoU gain obtained using fire area segmentation. Across the three departments, the \emph{fire-area} strategy consistently improves IoU over the grid baseline. Doubs shows the strongest uplift ($\Delta{=}+0.05$ across all scales); Ain benefits only at the mid- and high-thresholds, while Yvelines gains mainly at mid-scale.

\begin{table}[htbp]
	\centering
	\caption{Mean IoU gain $\Delta$ (fire area – grid) per department and lead time.}
	\footnotesize
	\setlength{\tabcolsep}{6pt}
	\begin{tabular}{lccc}
		\toprule
		\textbf{Department} & \textbf{$s=0.2$} & \textbf{$s=0.3$} & \textbf{$s=0.4$} \\
		\midrule
		\addlinespace[2pt]
		Ain                 & 0.00             & 0.04             & 0.04             \\
		Doubs               & 0.04             & 0.04             & 0.05             \\
		Yvelines            & 0.03             & 0.04             & 0.01             \\
		\bottomrule
	\end{tabular}
	\label{tab:mean_gain_firemen_leadtime}
\end{table}

\begin{table}[htbp]
	\centering
	\caption{IoU performance by department and scale on \textbf{Firefighter} dataset (0-day forecast).}
	\footnotesize
	\setlength{\tabcolsep}{4pt}
	\renewcommand{\arraystretch}{1.1}

	\begin{tabular}{lccc|ccc|ccc}
		\toprule
		      & \multicolumn{3}{c|}{\textbf{Ain}}
		      & \multicolumn{3}{c|}{\textbf{Doubs}}
		      & \multicolumn{3}{c}{\textbf{Yvelines}}                                                 \\
		Model & 0.2                                   & 0.3                   & 0.4
		      & 0.2                                   & 0.3                   & 0.4
		      & 0.2                                   & 0.3                   & 0.4                   \\
		\midrule

		\multicolumn{10}{c}{\textbf{Fire area}}                                                       \\
		LR
		      & 0.14                                  & 0.24                  & 0.34
		      & \textcolor{red}{0.12}                 & \textcolor{red}{0.19} & 0.25
		      & 0.17                                  & 0.26                  & 0.32                  \\

		XGBoost
		      & 0.15                                  & \textcolor{red}{0.26} & 0.33
		      & \textcolor{red}{0.12}                 & 0.18                  & \textcolor{red}{0.25}
		      & \textcolor{red}{0.18}                 & 0.26                  & 0.32                  \\

		CatBoost
		      & \textcolor{red}{0.16}                 & 0.25                  & \textcolor{red}{0.35}
		      & \textcolor{red}{0.12}                 & 0.18                  & 0.25
		      & 0.17                                  & 0.26                  & 0.34                  \\

		GRU
		      & 0.14                                  & 0.21                  & 0.33
		      & \textcolor{red}{0.12}                 & 0.17                  & 0.25
		      & 0.17                                  & 0.24                  & 0.33                  \\

		LSTM
		      & 0.14                                  & 0.24                  & 0.33
		      & \textcolor{red}{0.12}                 & 0.15                  & \textcolor{red}{0.26}
		      & 0.17                                  & 0.25                  & \textcolor{red}{0.35} \\

		MLP
		      & 0.14                                  & 0.25                  & 0.30
		      & \textcolor{red}{0.12}                 & \textcolor{red}{0.19} & 0.24
		      & 0.17                                  & \textcolor{red}{0.28} & 0.33                  \\

		\midrule
		\multicolumn{10}{c}{\textbf{Grid}}                                                            \\
		LR
		      & 0.13                                  & 0.21                  & 0.32
		      & 0.10                                  & 0.14                  & 0.20
		      & 0.13                                  & 0.20                  & 0.32                  \\

		XGBoost
		      & 0.14                                  & 0.20                  & 0.31
		      & 0.11                                  & 0.14                  & 0.22
		      & 0.15                                  & 0.23                  & 0.32                  \\

		CatBoost
		      & 0.14                                  & 0.22                  & 0.30
		      & 0.10                                  & 0.14                  & 0.20
		      & 0.15                                  & 0.22                  & 0.33                  \\

		GRU
		      & 0.13                                  & 0.21                  & 0.31
		      & 0.10                                  & 0.13                  & 0.20
		      & 0.15                                  & 0.21                  & 0.32                  \\

		LSTM
		      & 0.14                                  & 0.20                  & 0.29
		      & 0.09                                  & 0.13                  & 0.21
		      & 0.13                                  & 0.22                  & 0.33                  \\

		MLP
		      & 0.13                                  & 0.19                  & 0.23
		      & 0.10                                  & 0.14                  & 0.16
		      & 0.15                                  & 0.22                  & 0.31                  \\

		\bottomrule
	\end{tabular}
	\label{tab:performance_firemen}
\end{table}

\subsection{BDIFF}

Table \ref{tab:performance_bdiff} compares grid and fire-area segmentation performance in the \textbf{BDIFF} dataset. Table~\ref{tab:gain_bdiff_leadtimes} shows the average IoU gain obtained using fire-area segmentation. The \emph{fire-area} approach systematically outperforms the grid baseline. Bouches-du-Rhône gains most at the lower scale ($\Delta{=}+0.06$ at $s=0.2$) while Herault peaks at the mid-scale ($\Delta{=}+0.06$ at $s=0.3$).

\begin{table}[htbp]
	\centering
	\caption{IoU performance by department and scale on \textbf{BDIFF} dataset (0-day forecast). Red values indicate the best values for a given scale.}
	\footnotesize
	\setlength{\tabcolsep}{4pt}
	\renewcommand{\arraystretch}{1.1}

	\begin{tabular}{lccc|ccc}
		\toprule
		         & \multicolumn{3}{c|}{\textbf{Bouches-du-Rhone}} & \multicolumn{3}{c}{\textbf{Herault}}                                           \\
		Model    & 0.2                                            & 0.3                                  & 0.4                   & 0.2 & 0.3 & 0.4 \\
		\midrule

		\multicolumn{7}{c}{\textbf{Fire area}}                                                                                                     \\
		LR       & \textcolor{red}{0.26}                          & \textcolor{red}{0.30}                & 0.34
		         & 0.11                                           & \textcolor{red}{0.19}                & 0.23                                    \\
		XGBoost  & 0.25                                           & 0.29                                 & 0.34
		         & \textcolor{red}{0.13}                          & 0.17                                 & 0.22                                    \\
		CatBoost & 0.25                                           & \textcolor{red}{0.30}                & \textcolor{red}{0.35}
		         & 0.12                                           & 0.19                                 & \textcolor{red}{0.24}                   \\
		GRU      & 0.23                                           & 0.28                                 & 0.34
		         & 0.11                                           & 0.16                                 & 0.22                                    \\
		LSTM     & 0.23                                           & \textcolor{red}{0.30}                & 0.34
		         & 0.12                                           & 0.16                                 & 0.21                                    \\
		MLP      & 0.23                                           & 0.28                                 & 0.33
		         & 0.11                                           & 0.17                                 & 0.21                                    \\

		\midrule
		\multicolumn{7}{c}{\textbf{Grid}}                                                                                                          \\
		LR       & 0.17                                           & 0.29                                 & 0.32
		         & 0.065                                          & 0.12                                 & 0.20                                    \\
		XGBoost  & 0.20                                           & 0.25                                 & 0.32
		         & 0.086                                          & 0.11                                 & 0.20                                    \\
		CatBoost & 0.19                                           & 0.28                                 & 0.32
		         & 0.066                                          & 0.12                                 & 0.23                                    \\
		GRU      & 0.18                                           & 0.25                                 & 0.28
		         & 0.065                                          & 0.11                                 & 0.16                                    \\
		LSTM     & 0.16                                           & 0.26                                 & 0.29
		         & 0.060                                          & 0.11                                 & 0.18                                    \\
		MLP      & 0.19                                           & 0.25                                 & 0.29
		         & 0.10                                           & 0.13                                 & 0.23                                    \\

		\bottomrule
	\end{tabular}
	\label{tab:performance_bdiff}
\end{table}

\begin{table}[htbp]
	\centering
	\caption{Mean IoU gain $\Delta$ (fire area – grid) per department and lead time.}
	\footnotesize
	\setlength{\tabcolsep}{6pt}
	\begin{tabular}{lccc}
		\toprule
		\textbf{Department} & \textbf{$s=0.2$} & \textbf{$s=0.3$} & \textbf{$s=0.4$} \\
		\midrule
		\addlinespace[2pt]
		Bouches-du-Rhone    & 0.06             & 0.03             & 0.04             \\
		Herault             & 0.04             & 0.06             & 0.02             \\
		\bottomrule
	\end{tabular}
	\label{tab:gain_bdiff_leadtimes}
\end{table}

\begin{figure}[htbp]
	\centering
	\begin{tikzpicture}
		\begin{axis}[
				ybar,
				width=0.95\columnwidth,
				height=5cm,
				ylabel={IoU Gain (\%)},
				symbolic x coords={Ain, Doubs, Yvelines, B-du-R, Hérault},
				xtick=data,
				x tick label style={rotate=30, anchor=east, font=\footnotesize},
				ymin=0, ymax=8,
				bar width=4pt,
				legend style={at={(0.5,-0.25)}, anchor=north, legend columns=3, font=\scriptsize},
				enlarge x limits=0.15,
				nodes near coords,
				nodes near coords style={font=\tiny},
				every node near coord/.append style={yshift=2pt},
			]
			\addplot[fill=blue!70] coordinates {(Ain,0) (Doubs,4) (Yvelines,3) (B-du-R,6) (Hérault,4)};
			\addplot[fill=orange!70] coordinates {(Ain,4) (Doubs,4) (Yvelines,4) (B-du-R,3) (Hérault,6)};
			\addplot[fill=green!60] coordinates {(Ain,4) (Doubs,5) (Yvelines,1) (B-du-R,4) (Hérault,2)};
			\legend{$s{=}0.2$, $s{=}0.3$, $s{=}0.4$}
		\end{axis}
	\end{tikzpicture}
	\caption{Mean IoU improvement (fire-area over grid segmentation) across all departments and spatial scales. The medium scale ($s{=}0.3^{\circ}$) yields the most consistent gains.}
	\label{fig:iou_gains_summary}
\end{figure}
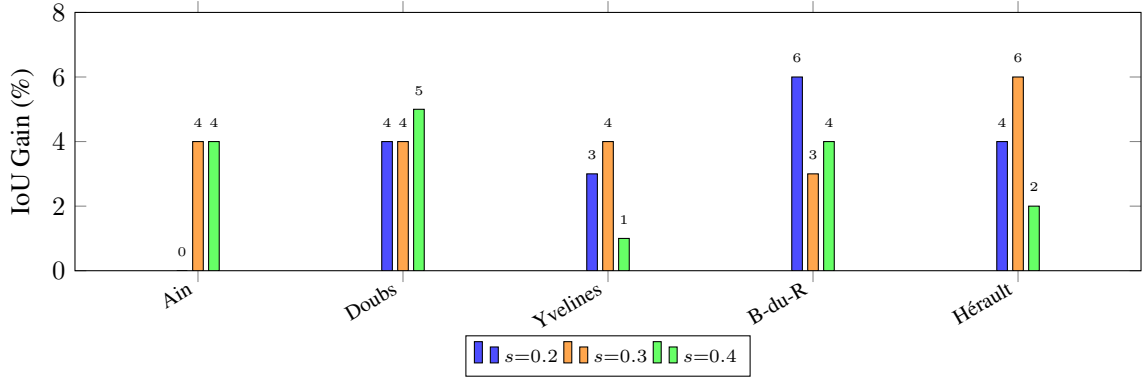

\begin{figure}[htbp]
	\centering
	\vskip -0.2in
	\includegraphics[scale=0.3]{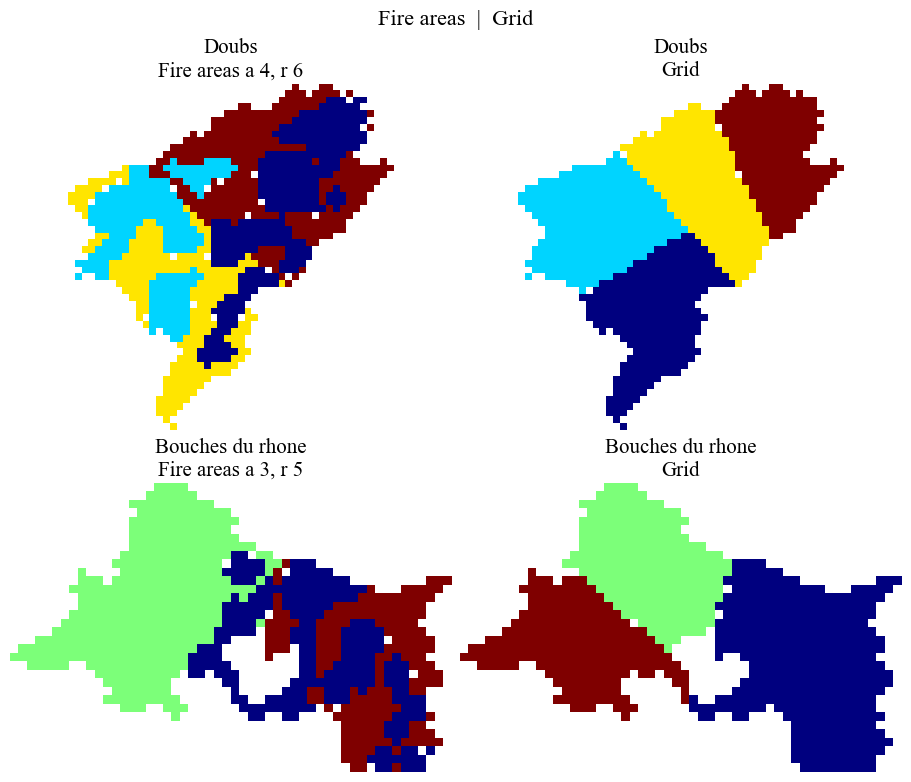}
	\caption{Comparison of watershed and grid segmentation with scale=0.3 and tolerance=0.3. Colors indicate unique IDs with no other meaning. $a=max\_dilations$, $\mathrm{r}=intensity\_levels$.}
	\label{fig:comparison}
	\vskip -0.2in
\end{figure}

We recommend fire-area segmentation as the primary method---a ``no-regrets'' improvement over grid-based discretization. The medium scale ($s=0.3^{\circ}$) yields the most robust cross-department performance (see Figure~\ref{fig:iou_gains_summary}). At finer scales ($s=0.2^{\circ}$), gains remain positive but metrics are more volatile; at coarser scales ($s=0.4^{\circ}$), the performance gap narrows. In no scenario does grid-based segmentation surpass fire-area segmentation. Figure~\ref{fig:comparison} compares both approaches for Doubs and Bouches-du-Rh\^one.

A reproducibility test using logistic regression revealed result variability of approximately 0.01 for both approaches, confirming that observed differences reflect genuine methodological improvements.

\subsection{Discussion}

The results highlight the key advantage of segmenting fire-prone areas prior to prediction: improved model learning. Our approach is complementary to supervised deep-learning methods (e.g., U-Net, ViT): it can serve as an unsupervised preprocessing step that defines semantically meaningful prediction units \emph{before} any downstream segmentation or classification model is applied. However, several limitations remain and will be addressed in future work.

\noindent\textbf{Parameter coupling across departments.} The current parameter optimization assumes that the algorithm's optimal parameters are identical for every department, which is incorrect---for example, Bouches-du-Rh\^one requires 2 dilations, whereas H\'erault requires 3. To quantify this coupling effect, we measured the performance degradation when using a single configuration (optimized on Bouches-du-Rh\^one) across all departments: mean IoU dropped by approximately 0.02--0.04 for H\'erault and Ain, confirming that departmental heterogeneity introduces non-negligible variance. Detailed per-department configurations and their cross-application effects are provided in the supplementary materials. Two avenues could mitigate this coupling:
\begin{enumerate}
	\item \textbf{Exhaustive cross-configuration experiment.}
	      Rather than applying a single segmentation to all regions, a separate experiment can be run for \emph{every possible combination} of departmental configurations at the same scale. Although this increases computational cost, the procedure systematically explores the entire space of segmentation choices. Depending on the model and hardware, training could easily require several weeks.

	\item \textbf{Department-specific models.} Training one model per department would achieve the same isolation more economically; however, the smaller number of fire events per dataset could hurt generalization and degrade overall performance, potentially leading to suboptimal segmentation.
\end{enumerate}

\noindent\textbf{Scalability considerations.} The segmentation algorithm has modest computational requirements. On our hardware (AMD EPYC 7513, 256\,GB RAM), processing a single department at 2\,km resolution ($\approx$100$\times$100 pixels) takes less than 10 seconds per configuration; the full grid-search over 25 configurations completes in under 5 minutes per department. Memory footprint remains below 2\,GB even for the largest departments. For national-scale deployment, the algorithm can be applied department-by-department in a fully parallelizable fashion. At finer resolutions ($<$1\,km), memory and time costs would increase quadratically with pixel count, suggesting that tiled processing or downsampled density estimation would be required.

\noindent\textbf{Cross-boundary zones.} We acknowledge that the current experiments do not handle zones that straddle departmental boundaries, a limitation arising from three considerations: (i)~French fire services are organized at the departmental level, so operational risk assessments must be reported by department; (ii)~aggregating by department reduces the volume of data that must be stored in memory; and (iii)~we lack access to a complete, nationwide record of fire-brigade interventions.

\noindent\textbf{Data-scarce regions.} When a region has very limited wildfire experience or when no historical fire records are available, the segmentation process becomes challenging with the current algorithm due to the absence of ignition points. To address this issue, we consider two strategies. First, a spatial model (e.g., CNN, U-Net, SwinUNet) can be trained to learn the segmentation produced by the algorithm on data-rich regions and then used to infer a segmentation for previously unseen or data-scarce regions. Second, an upstream fire-susceptibility map could be constructed and used as input to the segmentation algorithm.

\section{Conclusion}
In this study, we demonstrate that conventional grid-based segmentation is poorly suited to wildfire prediction because it introduces spatial noise that weakens the model's ability to learn ignition patterns. To address this limitation, we propose a fire-zone segmentation method that defines prediction units directly from historical fire distributions.

Evaluated across six French departments, six forecasting models, three spatial scales, and multiple lead times, this data-driven segmentation consistently outperforms the static grid approach with mean IoU improvements of +3--6\%. The method requires no labeled segmentation data, runs in under 10 seconds per configuration, and integrates seamlessly into existing prediction pipelines.

The parameter search further indicates that no single configuration is optimal for all departments, highlighting the importance of region-specific tuning. For operational deployment, we recommend the medium scale ($s=0.3^{\circ}$) as a robust default. Future work may explore exhaustive cross-configuration testing, transfer learning from data-rich to data-scarce regions, or coupling the segmentation with downstream deep-learning models for end-to-end optimization.

\bibliographystyle{plain}  
\bibliography{main}

\end{document}